\documentclass[11pt,a4paper]{article}
\usepackage[hyperref]{conll2018}
\usepackage{times}
\usepackage{latexsym}
\usepackage[pdftex]{graphicx} 

\usepackage{booktabs} % For formal tables
\usepackage{flushend} %this is to level the references
\usepackage{multirow}
\usepackage{subfigure}
\usepackage{enumitem}

\usepackage{url}

\aclfinalcopy % Uncomment this line for the final submission

\title{Modular Approach to Machine Reading Comprehension: Mixture of Task-Aware Experts}

\author{%
    Anirudha Rayasam \\
  \small Language Technologies Institute \\
  \small School of Computer Science \\
  \small Carnegie Mellon University \\
  {\tt \small arayasam@andrew.cmu.edu} \\\And
  Anusha Kamath \\
  \small Language Technologies Institute \\
  \small School of Computer Science \\
  \small Carnegie Mellon University \\
  {\tt \small akamath1@andrew.cmu.edu} \\\And 
  Gabriel Bayomi Tinoco Kalejaiye \\
  \small Language Technologies Institute \\
  \small School of Computer Science \\
  \small Carnegie Mellon University \\
  {\tt \small gbayomi@andrew.cmu.edu} \\}

\date{}

\begin{document}
\maketitle

\begin{abstract}

  In this work we present a Mixture of Task-Aware Experts Network for Machine Reading Comprehension on a relatively small dataset. We particularly focus on the issue of commonsense learning, enforcing the common ground knowledge by specifically training different expert networks to capture different kinds of relationships between each passage, question and choice triplet. Moreover, we take inspiration on the recent advancements of multi-task and transfer learning by training each network a relevant focused task. By making the mixture-of-networks aware of a specific goal by enforcing a task and a relationship, we achieve state-of-the-art results and reduce over-fitting.
  
\end{abstract}

\section{Introduction}
Teaching a computer to read and comprehend human languages is a challenging task for machines as this requires the understanding of natural languages and the ability to reason over various clues. The task of Machine Reading Comprehension (MRC) is a useful benchmark to demonstrate natural language understanding. In recent years, several datasets have been created to focus on answering questions as a way to evaluate machine comprehension. The machine is first presented with a piece of text such as a news article or a story and is then expected to answer one or multiple questions related to the text. One of the big challenges of the field is to provide a system that is able to infer relationships that are not entirely based on the passage: reaching commonsense.

Powerful approaches have been explored to solve this issue by applying attention mechanisms over passage, choice and answer. However, by trying to understand the three different aspects concomitantly, they usually fail to capture the distinct commonsense relationship between question-choice or passage-choice pairs. 

We propose the expert networks QC (question-choice) and PC(passage-choice) which specifically are trained to learn the different structures necessary to answer the questions. Our goal is to achieve commonsense knowledge by directly enforcing the network to build question-choice and passage-choice relationships. For a human being, regardless of the passage, it should be clear that ``fireworks" is a more likely answer than ``water" for the question ``how did the fire start?" unless the passage clearly states the contrary. Additionally, inspired by the efficacy of multi-task learning techniques, we propose a Task-Aware Expert Training, where each network is trained on  a different but relevant task in order to improve their overall inference capabilities.  

\section{Related work}

A lot of the work in Neural Machine comprehension focuses on how to extract the required information from the given passage. Recent approaches have had enormous success, for instance: R-net \cite{rnet} and Match-LSTM \cite{match_lstm}. However this might not be a good solution for MRC when there is a need of generating additional text not included in the passage or the question, augmenting information in multiple passage spans and the question as and when required. The work from \cite{memnets},  \cite{snet}, \cite{attoveratt} are examples of this change of paradigm. An interesting adaptation  involves using single or multiple turns of reasoning to effectively exploit the relation among queries, documents, and answers. \cite{tris}, \cite{hermann}, \cite{reasonnet}, \cite{multistrategy}, \cite{gupta2019amazonqa}, \cite{anirudha2014genetic} and \cite{larionov2018tartan} are great examples of recent relevant strategies. Some more relevant work is available in \cite{anirudhaonline, anirudhasoft}

One intuitive line of work in Machine comprehension uses common sense knowledge along with the comprehension text to generate answers. Common-sense, knowledge increases the accuracy of machine comprehension systems. The challenge is to find a way to include this additional data and improve the system's performance. There are many possible common-sense knowledge sources. Generally, script knowledge which is sequences of events that describe typical human actions in an everyday situations is used.  

The work from \cite{reason} shows how a multi-knowledge reasoning method, which explores heterogeneous knowledge relationships, can be powerful for commonsense MRC. This approach is achieved by combining different kinds of structures for knowledge: narrative knowledge, entity semantic knowledge and sentiment coherent knowledge. By using data mining techniques, they provide a model with cost-based inference rules as an encoding mechanism for knowledge. Later, they are able to produce a multi-knowledge reasoning model that has the ability to select which inference rule to use for each context. 

Another interesting approach comes from \cite{ganmrc} where the authors proposed the usage of Conditional Generative Adversarial Network (CGAN) to tackle the problem of insufficient data for reading comprehension tasks by generating additional fake sentences and the proposed generator is conditioned by the referred context, achieving state-of-the-art results. 

The work by \cite{sota} assesses how a Three-Way Attentive Network (TriAN) with the inclusion of commonsense knowledge benefits multiple choice reading comprehension. The combination of attention mechanisms have shown to strongly improve performance for reading comprehension. In addition to that, commonsense knowledge can help in inferring nontrivial implicit events within the comprehension passage. The work by \cite{commonsense} focuses on reasoning with heterogeneous commonsense knowledge. They use three kinds of commonsense knowledge, causal relations, semantic relations like co-reference,associative relations and lastly sentiment knowledge - sentiment coherence (positivity and negativity) between two elements. In human reasoning process, not all inference rules have the same possibility to be applied, because the more reasonable inference will be proposed more likely. They use attention to weigh the inferences based on the nature of the rule and the given context. Their attention mechanism models the possibility that an inference rule is applied during the inference from a premise document to a hypothesis by considering the relatedness between elements and knowledge category, as well as the relatedness between two elements. They answer the comprehension task by summarizing over all valid inference rules.

Although Mixture of Experts models are widely used for NLP tasks, it is still underused for Machine Reading Comprehension tasks. Recent work includes the Language Model paper from \cite{mixlstm} which introduces an LSTM-based mixture method for the dynamic integration of a group of word prediction experts in order to achieve conditional language model which excels simultaneously at several subtasks. Moreover, the work from \cite{dcn} also includes a sparse mixture of experts
layer for a Question-Answering task, which is inherited from the previous work of \cite{outrage} on a Sparsely-Gated Mixture-of-Experts Layer. The success of the aforementioned approaches show the optimism about introducing Mixture of Experts Deep Learning for Machine Reading Comprehension.

\section{Model description}
We experiment with a mixture of experts model to tackle the task of commonsense machine comprehension. The model is inspired from analyzing the errors made by a triple attention network from \cite{trian} which achieves state-of-the-art results by using a Three-Way Attentive Network (TriAN) with the inclusion of commonsense knowledge in the form of relational embeddings obtained from the ConceptNet knowledge graph which is a large-scale graph of commonsense knowledge consisting of over 21 million edges and 8 million nodes.

A training example in the commonsense comprehension task consists of a passage (P), question (Q), answer (A) and label y which is 0 or 1. P, Q and A are all sequences of words. For a word $P_i$ in the given passage, the input representation of $P_i$ is the concatenation of several vectors: Pre-trained GloVe embeddings, Part-of-speech (POS) embedding, named-entity embedding (NE), ConceptNet Relation embedding and handcrafted features - term frequency and co-occurence. The 10 dimensional POS, 12 dimensional NE and 10 dimensional relation embeddings are randomly initialized. The relation in the relation embedding is determined by querying ConceptNet.

The passage, question and choices are passed through the embedding layers and bi-RNNs to obtain the passage,question and choice embedding vectors. These vectors are further passed through multiple attention layers to capture the interaction between each pair. The understanding of interactions between passage, question and answers are fundamental to machine reading comprehension tasks. Additionally, different triples of this information might need to focus on different parts of each other. The TriAN (triple attention) architecture intends to address this issue by using Passage-Question, Question-Choice and Passage-Choice attentions in the architecture. Due to the relatively small size of training data in the SemEval task, TriAN uses word-level attention and consists of only one layer of LSTM to avoid over-fitting. They use standard sequence attention to obtain a question-aware passage representation, passage-aware answer representation and a question-aware answer representation. The passage feature embeddings are concatenated with the question aware passage vector and are then passed through a bidirectional RNN. Similarly, the sequence attention weighted choice and question vectors are concatenated with choice feature embeddings and question feature embeddings and are passed through a Bi-RNN. Thus obtained passage,choice and question embeddings are further passed through a self-attention layer and the output classification probability is obtained as a bilinear function over the self attention weighted representations. Adopting the context-aware representations proposed in this work we propose extensions motivated by the analysis of the failure case of the TriAN model.

\begin{figure*}[t]
	\centering
   \includegraphics[width=0.85\textwidth]{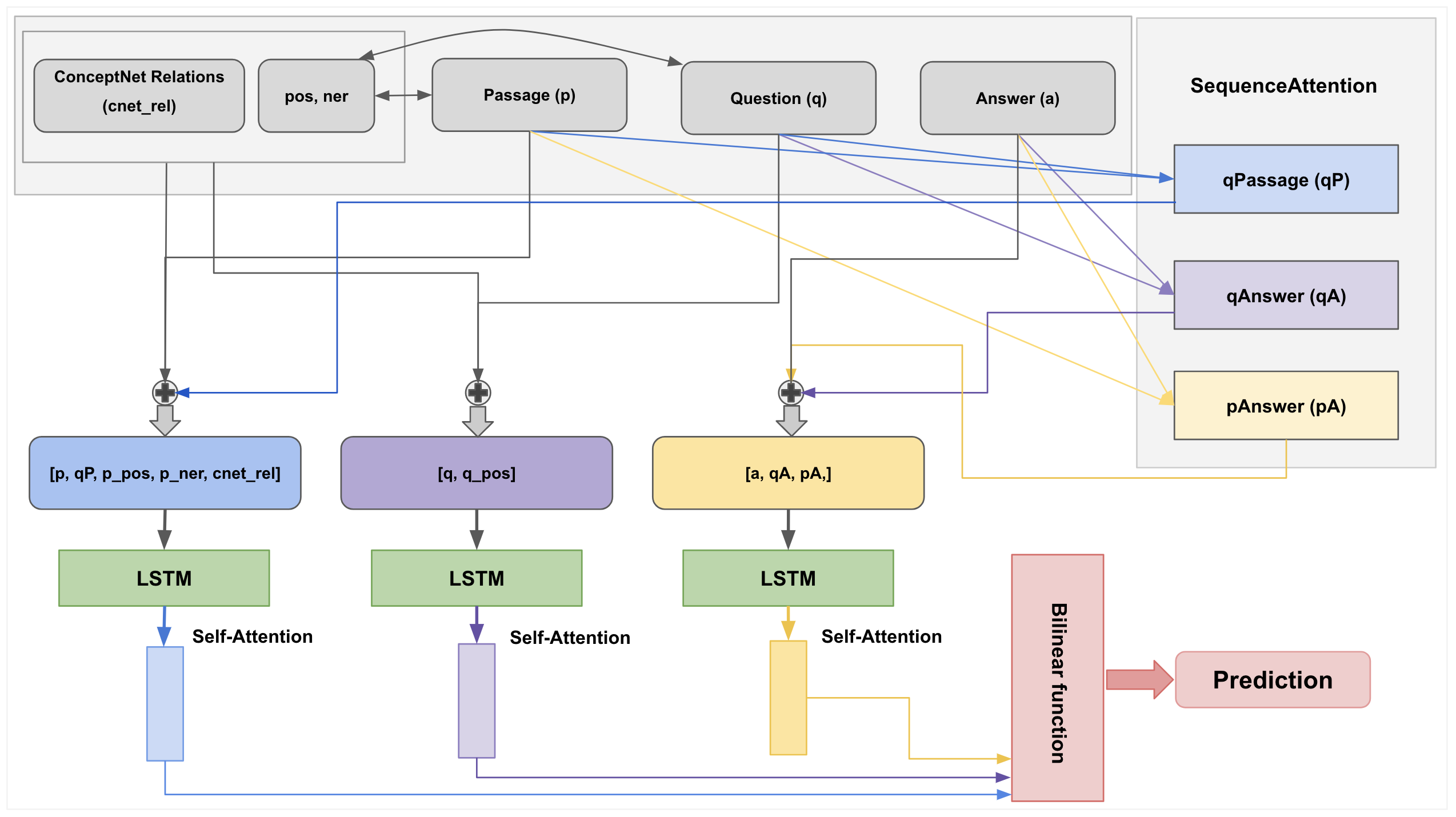}
   \caption{Baseline Architecture : Three-way Attentive Networks}
   \label{fig: accuracy curve}
\end{figure*}

\subsection{Motivation for Auxiliary Methods from Error Analysis}

The challenge in the Sem-Eval 2018 Task 11 is to improve the performance on comprehension tasks by increasing the commonsense capabilities of the neural network. The baseline paper approached this by using concept-net to obtain enhanced passage and choice representations by means of adding relational embeddings based on the 27 relations in the ConceptNet Knowledge graph. We found that despite the usage of a commonsense KB, most of the failures of the network fell into the category of lack of commonsense knowledge. A majority of the errors could be classified into 3 sets. First, the network didn't do well when some information was implicitly conveyed in the passage. Second, the network failed to answer questions that required understanding and connecting a sequence of events. Third, a very clear deficiency of the baseline model is that it fails to capture choices that are obviously false in any context. It means that on a few instances the passage actually doesn't add much information and because of their large size it hurts the overall performance by adding content which is not useful. Finally, another important problem is overfitting: the complex structure helps on capturing context between passage, choice and answer; but the model performance is degraded for the development set. Therefore, we decided to explicit model question-choice relationships in order to capture obviously unlikely answers and passage-choice relationships to understand and connect a sequence of events from the passage. Moreover, we tackle the over-fitting problem by indirect data augmentation by pre-training the relationships with textual entailment and story completion.

\subsection{Task-Aware Expert Training}

The error analysis and assumptions aforementioned motivated us to pre-train the network on tasks which could create an inductive bias in the network in terms of being able to infer implicit content in the passage and in terms of understanding sequentially connected events.\\

An important task to improve the performance of the TriAn is to capture the question-choice and passage-choice relationships cited above. Our main contribution is the development of auxiliary networks. Basically, we propose a combination of a Question-Choice Network (ignoring the passage information, henceforth called QCN); a Passage-Choice Network (ignoring the question, henceforth called PCN); and the actual baseline Passage-Question-Choice Network (henceforth called PQCN). Another problem that the current architecture faced was with regards to over-fitting. The Sem-Eval dataset for Machine comprehension with commonsense knowledge is relatively smaller in comparison to most comprehension datasets and we found that the network was very easily overfitting to this dataset despite usage of standard techniques like dropout regularization. We approached this by pre-training the network on auxiliary tasks like entailment (for QCN) and story cloze completion (for PCN). The motivation was that tasks like entailment and story completion require commonsense skills like interpreting implicit information and understanding and connecting sequence of events within a passage.   \\

We observed that we could obtain results of around 77.5\% on the development set by just using the question-choice pairs and discarding the passages in the comprehension using QCN. Similarly, we could obtain a performance of 76.5\% on the development set using only the passage-choice pairs using PCN. Moreover, we observed that, although the sub-networks (i.e QCN and PCN) performed worse than the Passage-Question-Choice network (PQCN), the networks tended to make different kinds of errors mostly because the input information to these networks were different.\\

We tried two approaches of integrating these three streams in our architecture. Firstly, we tried a multi-stream mixture of experts approach where we used shared network layers across the different streams. For instance, the question RNN was tied between the QC net and the PQC net and the passage RNN was tied between the PC net and the PQC net. In addition to this, embedding layers and attention layers were tied across the streams. Tying the layers however forced the individual networks to make similar mistakes and we did not see any boost in the performance of this network on the development set. \\

To bump up the performance of the individual streams, we pre-trained the QC net on entailment (SNLI dataset) and we trained the PC net on the story cloze dataset (ROC stories). By analyzing the mistakes made by the baseline architecture we felt that both these skills were lacking to some extent in the current architecture. After pre-training the individual streams the performance of the question-choice net improved to 81.2\% from 77.5\%  and the performance of the passage-choice net improved to 77.8\% from 76.2\%. We used the PCN, QCN and PQCN in a multi-stream mixture architecture where each of the streams can be treated as individual experts making this architecture similar to a mixture of experts setup. The Figure 2 \label{fig:multi} shows the multi-stream architecture. \\

In the training phase, we equally weigh the predictions from both the streams, however when testing we weigh the prediction from each network by a confidence score. The design of the confidence score was motivated from error analysis where we observed that the error cases have very similar probabilities associated with both the choices i.e either both the choice are predicted to be the wrong choices (a very low predicted probability assigned to both choices in the range 0-0.2) or both the choices are predicted to be the right choices (a very high predicted probability assigned to both choices in the range of 0.8-1). This happens because we treat the problem as a binary classification problem where a choice is right or wrong, given the passage and the question irrespective of the other choice. To tackle this to some extent we weigh each network's prediction by the difference in the output probabilities associated with each choice. For instance if QCN has an output prediction of [Choice 1: 0.92, Choice 2: 0.85] - the weight for the QCN stream would be 0.07 and meanwhile if the prediction of the PQC network is [Choice 1: 0.2 Choice 2: 0.9], the weight given to the PQC network will be 0.7. In most cases when the individual streams select the same choice as the correct answer the weighted sum of the probabilities gives the same prediction as using individual streams. However, when the networks disagree a higher weight will be given to the network with a higher confidence score.\\

We also tried using a hard choice between networks. That is, we use the prediction from the network which has a higher confidence score as the output prediction. However, empirically, we found the weighted sum prediction performed slightly better than using a hard prediction. In the end, PQN didn't add as much value as initially predicted, committing very similar mistakes in comparison with PQCN. Therefore, our final multi-stream architecture contains only QCN and PQCN which are the main drivers of our result.

\begin{figure}[t]
	\centering
   \includegraphics[width=0.7\textwidth]{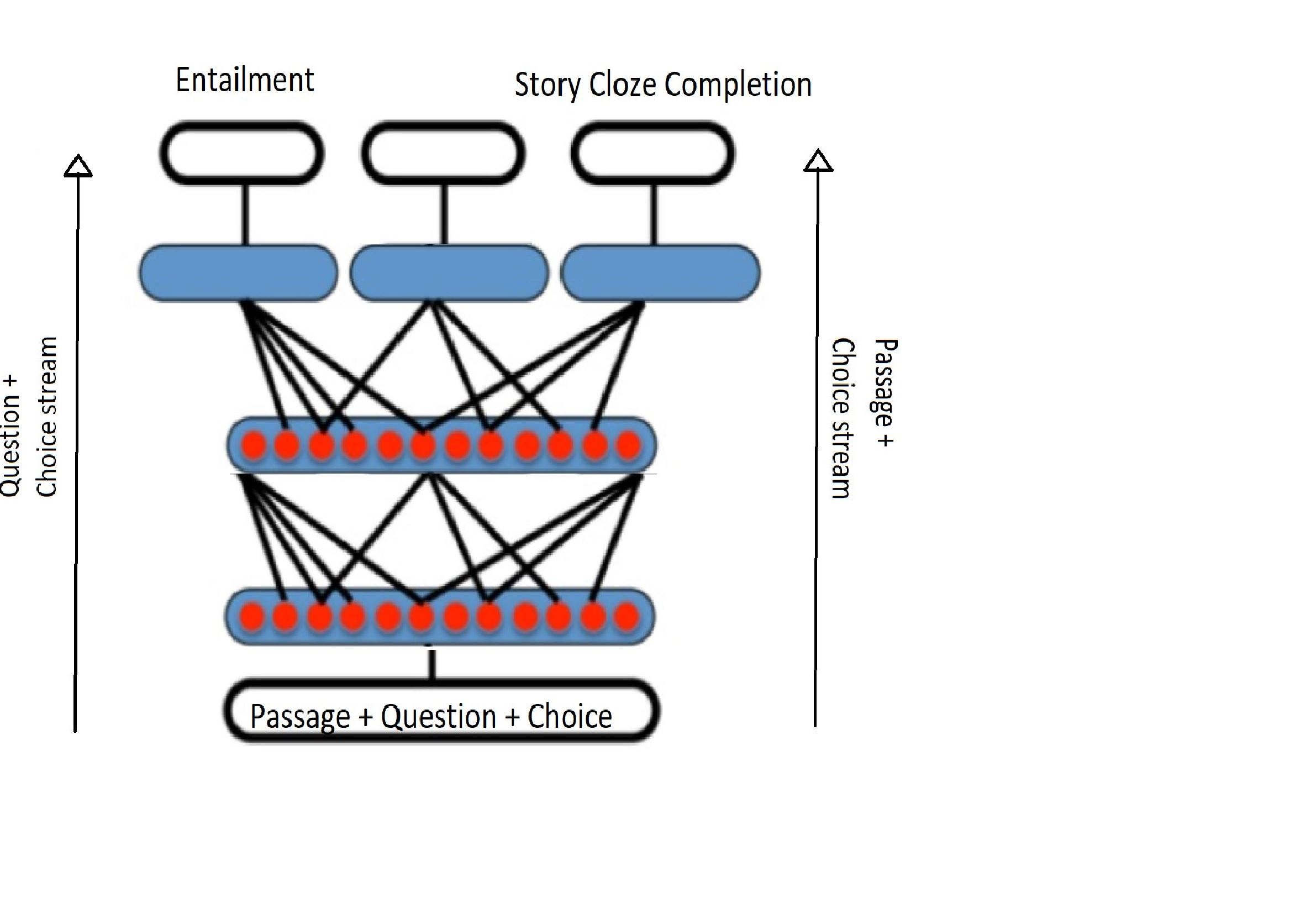}
   \caption{Proposed mixture of experts architecture}
   \label{fig:multi}
\end{figure}

% \usepackage[normalem]{ulem}
\begin{table*}[]
\tiny
\centering
\caption{Result analysis of the experts in the mixture-of-experts}
\label{error_comp}
\begin{tabular}{|l|l|l|}
\hline
% \vspace{1mm}
\textbf{SAMPLE (Excerpt)}                                                                                                                                                                  &  \textbf{QC-Net} & \textbf{PC-Net} \\ \hline
\begin{tabular}[c]{@{}l@{}}
My family and I decided that the evening was beautiful so we wanted to have a 
bonfire ., First , my husband went to our shed and \\ gathered some dry wood . I placed 
camp chairs around our fire pit .,Then my husband placed the dry wood in a pyramid
shape \\ inside the fire pit . He then set some small kindling ablaze using a long lighter
.,Once the kindling reached the dry wood , it set \\ the wood on fire .... I put
away our camp chairs . My husband made sure the fire was out by dousing it with some
water and we  \\ went inside to bed ., 

\vspace{1mm}

\\ Where did they light the fire ? \\ 
Their fire pit \\ 
in the house
\vspace{1mm}
\\ 

\textbf{Reason : The passage has fire-pit mentioned twice and passage choice attention helps in giving a higher probability to this} \\
\textbf{choice. When the question is included, probably the 'Where' in the question directs the answer towards house more than the fire-pit. }
\end{tabular}                                             

                                                                                                                                                                                                                                                                                                                                                                                                                                                                                                                                                                                                                                                                                                                                                                   &  Wrong  &  Correct  \\ \hline
\begin{tabular}[c]{@{}l@{}}
I needed to clean up my flat . I had to get my broom and vacuum and all the cleaners
I would need . I started by going around and \\ picking up any garbage I seem , like
used candy wrappers and old bottles of water . I threw them away . I went around and 
picked up \\ and dishes that were out and put them in the sink then washed them . I used
my broom to sweep up all the dust and dirt off the floors \\ the hard floors , and the
vacuum to clean up the floors that had rugs and carpets .... The t.v . , the 
tables and counters and everything \\ else that was a hard surface , I used the rag to
dust . My flat looked very nice when it was clean .

\vspace{1mm}

\\ Why did the flat needed cleaning ? 
\\ It was a big mess . \\ 
The dishes were washed at the sink .\\

\textbf{Reason : This question is far easier answered with just the question} \\
\textbf{than introducing the passage which also has the words 'dishes in the sink'.}
\end{tabular}

&  Correct  &  Wrong  \\ \hline

\begin{tabular}[c]{@{}l@{}}
Last night , my friends and I wanted to make a bonfire . So , we made sure that we
had some fire wood . We went to the back of her \\ yard and gathered up some dry wood .
My friend reminded us that the wood had to be dry , or else it would not burn very
well .... \\ As the fire continued to burn , we added more wood or newspaper to keep
the fire burning . We all had a fun time sitting around \\ the fire and talking .
\vspace{1mm}

\\ What kind of wood did they use to build the bonfire ? \\ 
Dry wood . \\ 
Wet wood . \\

\textbf{Reason : The passage has 'dry-wood' and 'dry' mentioned over and over which is captured by passage choice attention.} \\ 
\textbf{However given the question (introducing question choice attention) both PQC and QC networks get this wrong} \\

\end{tabular}                                                                                                                                                                                                                                                                                                                                                                                                                                                                                                                                                                                                                                                                                                                                                                                                                                                                                                                                       & Wrong   &  Correct  \\ \hline

\begin{tabular}[c]{@{}l@{}}One day I was super hungry and craving a delicious cheese
, spinach , and tomato pizza ! I decide to alleviate my hunger and my \\ craving by 
ordering a pizza from Domino 's . I called Domino 's to leave my order , and told them
exactly what I wanted in a size \\ medium . I did not order anything else when they asked 
me if I would like anything in addition ! They then asked me if I would \\ like delivery 
or pick up . I wanted to spend the evening relaxing after a long day and just watch 
movies with my pizza , so I opted \\ for delivery .... It was a successful evening and the
pizza was scrumptious !
\vspace{1mm}

\\ Who is cooking the pizza ? \\ 
Pizza Hut \\ 
Domino 's Pizza\\
\textbf{Reason : The PC-net gets this right because 'Dominos' in mentioned in the passage multiple times. However the QC-net}\\
\textbf{ends up giving similar scores to both the choices and ends up choosing the wrong option.} \\

\end{tabular}                                                                                                                                                                                                                                                                                                                                                                                                                                                                                                                   &  Wrong  & Correct   \\ \hline

\end{tabular}
\end{table*}

\section{\textbf{Model Analysis and Conclusions}}
The Multi-Stream Mixture of Experts (with only QCN and PQCN layers) network obtains an accuracy of 84.3\% on the development set which is a 0.46\% improvement over the baseline architecture which achieved an accuracy of 83.84\% after pre-training the model on RACE dataset. The QC stream in our architecture was pre-trained on entailment data (SNLI dataset), however the PQCN stream was trained on the SemEval dataset from scratch. As we didn't make any random-seed-ensemble methods, we didn't include them as a comparison baseline. Figure 3 shows the development set learning curves for the two architectures.
\\

Table \ref{error_comp} shows the comparison of the errors made by different streams PQC, QC, PC. We list our observations below. Below we summarize the error analysis on the baseline architecture which corresponds to the PQC stream in our current model. 

\begin{table}[h]
\centering
\small
\caption{Ablation studies performed on the model by excluding some of the features}

\label{ablation}
\begin{tabular}{l|c}
\hline
\textbf{Features}    & \textbf{Dev Accuracy} \\ \hline
w/o POS                  & 82.1\%               \\
w/o NER                  & 82.5\%               \\
w/o Handcrafted Features & 82.5\%               \\
w/o ConceptNet           & 81.8\%               \\
w/o GloVe embeddings     & 77.5\%               \\
\textbf{Baseline including all features} & \textbf{82.7}\%        \\
\textbf{Baseline Model with pre-training}   & \textbf{83.84}\%	\\
\hline
QC stream                & 77.5\% \\ 
PC stream				 & 76.2\% \\
QC stream (pre-trained on entailment)               & 81.2\% \\
PC stream (pre-trained on story-cloze)               & 77.8\% \\
\textbf{Mixture of experts (PQCN \& QCN) }   & \textbf{83.2}\%	\\
\textbf{Mixture of experts (after pre-training) }   & \textbf{84.5}\%	\\
\hline
\end{tabular}
\end{table}

\begin{figure}[ht]
	\centering
   \includegraphics[width=0.5\textwidth]{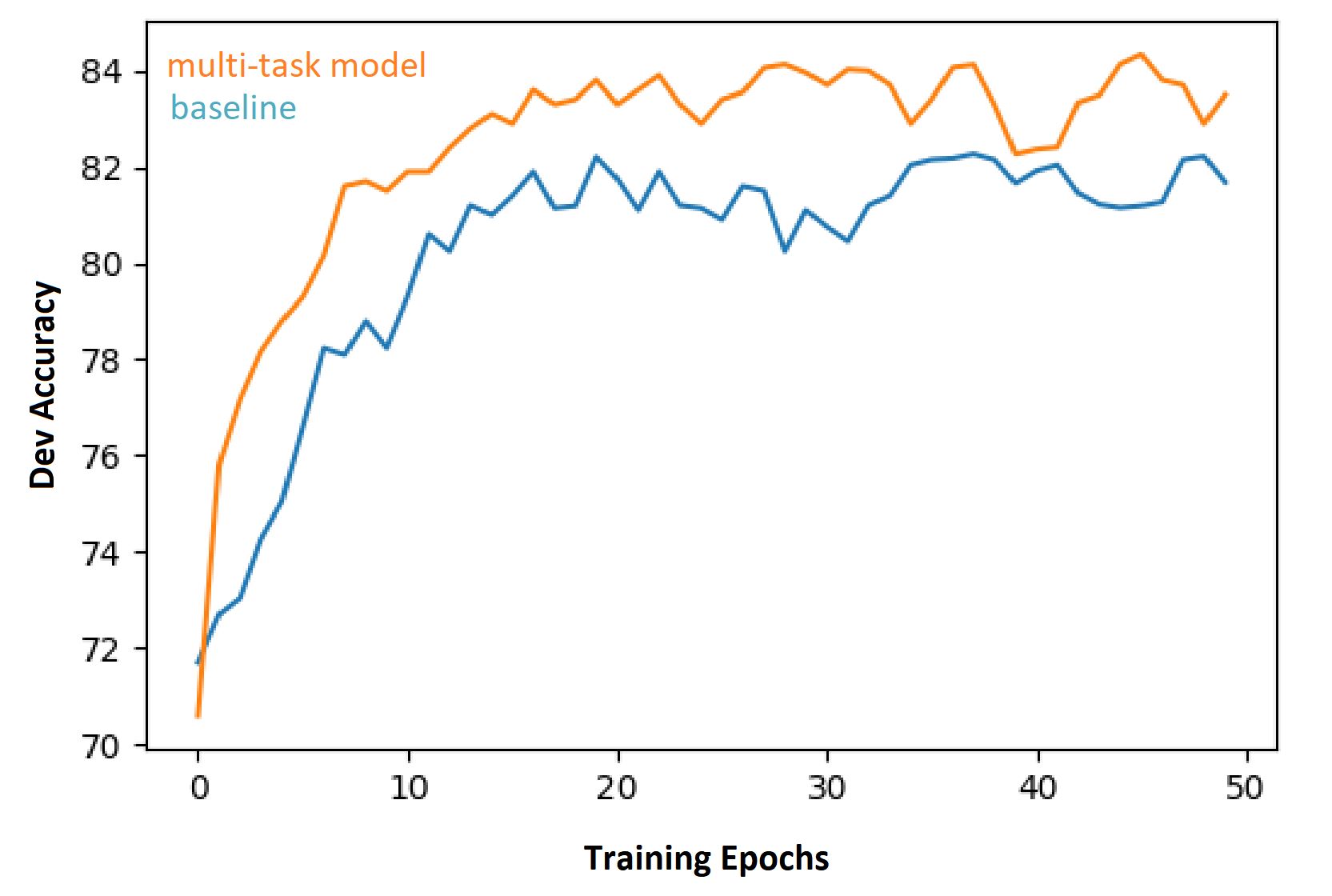}
   \caption{Dev accuracy vs training epochs}
   \label{fig: accuracy curve}
\end{figure}

\begin{itemize}[leftmargin=-.005in]
\item Many failure questions require understanding of a sequence of events. An expert on story completion task could help with answering these questions. We pre-train the passage-choice network on the story completion dataset and use it as one of the experts in a mixture of experts.
% \vspace{3mm}

\item In many cases, the choice is entailed just from the question \& the QC network can answer this accurately without any further context from the passage. We can see instances from the table where the PQC network fails in selecting the right choice. However the QC network when used in the mixture of experts helps remedy this.
% \vspace{3mm}

\item In cases where the answer can be answered without further context from the passage the problem is similar to checking entailment between  the question and choice sentences. We can see from the results table that the accuracy further improves if we pre-train the QC network on entailment and use it as one of the experts in the mixture. 
% \vspace{3mm}

\item As can be seen from the graph, the baseline architecture is very prone to over-fitting. Pre-training with auxiliary tasks like story completion and entailment helps in remedying this. In addition to this, these tasks might also be helping in improving the commonsense reasoning abilities of the network.
% \vspace{3mm}

\item The baseline (PQC) network also fails to understand which subject is related to what predicate and this leads to many errors as can be seen from the table. Using a dependency parse of the passage sentences as an additional input should help with this. However in some instances we found that after pre-training on auxiliary tasks the network performed better on questions that required understanding the underlying dependencies. Intuitively, the auxiliary tasks themselves require learning the dependencies in the sentence and this helps solve the problem to some extent. However this still remains as a problem in the current architecture.

\item The multi-stream architecture still lacks commonsense reasoning abilities and finds it especially hard to answer Yes/No questions that make it hard to place the choice in the context of the passage. Yes/No questions make up around 35\% of the dataset and contribute to 42\% of the errors made by the current network.

\item Pre-training on entailment surely helped in answering questions whose choices are implied just from the passage. But the network fails to capture choices that are entailed from the passage.

\item The passages are too long and the Passage Bi-LSTM is not able to capture many long term dependencies required in the comprehension task. Capturing co-reference and dependencies in the passage is hard to achieve with this architecture.

\item The different kinds of errors seen in the current architecture can be classified into :
\begin{itemize}[leftmargin=10.25pt]
\item Yes/No : Yes no questions tend to be harder for the network to answer because of lack of expert. \\
\item Paraphrase : When the choice options uses words that are semantically equivalent to the words in the passage. \\
\item External knowledge : Some questions require excess information to be answered correctly. This information at times cannot even be considered to be commonsense \\
\item Dependency Parse : A majority of the errors fall under this category. The Bi-LSTM does not capture many long distance dependencies which would be easier captured if there was a direct link via the parse tree. \\
\item Co-reference : Some errors, where an individual performs a sequence of events and in the later events the individual is referred by a pronoun. \\
\end{itemize}

Please find the different errors and corresponding classes in Table \ref{current_errors}
% \vspace{3mm}
\end{itemize}

% \usepackage[normalem]{ulem}
\begin{table*}[]
\tiny
\small
\caption{Result analysis of the mixture-of-expert model to propose further extensions}
\label{current_errors}
\begin{tabular}{|l|c|}
\hline
\textbf{ERROR SAMPLE (Excerpt)}                                                                                                                                                                                                                                                                                             & \textbf{ERROR CLASSES} \\ \hline
\begin{tabular}[c]{@{}l@{}}My family and I decided that the evening was beautiful so we wanted to have a bonfire . First , \\ my husband went to our shed and gathered some dry wood . I placed camp chairs around our \\ fire pit . Then my husband placed the dry wood in a pyramid shape inside the fire pit . He then \\ the marshmallows were golden brown , we placed them between two graham crackers with .... \\ utiful fire . When we finished , I put away our camp chairs . My husband made sure the fire was \\ out by dousing it with some water and we went inside to bed . 
\vspace{1mm}
\\ What did they do at the bonfire ?\\ Talk and cook smores .\\ 
They were dancing around . \end{tabular}                           
&        Paraphrase            \\ \hline
\begin{tabular}[c]{@{}l@{}}
I like to stay in touch with my family and friends . Usually when we are busy we stay in touch \\ by texting but we like to hear each others voices so we often call each other . My friends and \\ brother like to use our cell phones . I give them all a special ring tone so I know who it is right \\ .... I take the receiver off the cradle its sitting on and hold it to my ear and say " Hello . " The person \\ on the other end will have to introduce themselves , because I won't know who has called . But \\ usually I am lucky and its someone I want to talk to , so its a fun surprise .
\vspace{1mm}
\\ How did they answer the phone ?\\ By swiping the red phone icon\\ Swiping the green phone icon .\end{tabular} &   External\_knowledge                     \\ \hline
\begin{tabular}[c]{@{}l@{}}It was a Monday evening in a warm July heat wave . My wife and I had just completed a particularly \\ tasty meal that I had prepared for the both of us . We were both very full and sat in momentary \\ silence as we pondered the ... aware , napkins and drink glasses from the table and put them in the \\ down the table - top . After the plates , silverware , napkins and glasses were clear , she sprayed the \\ wood cleaner across the surface of the table . She took a white rag and wiped the table top until she \\ was satisfied that it was clean .
\vspace{1mm}
\\ Did the table have a table cloth ?\\ Yes\\ No .\end{tabular}                                                                                                                                                                                       &    Yes/No                    \\ \hline

\begin{tabular}[c]{@{}l@{}}I recently got a haircut . I wasn't happy with my previous haircut , so I wanted to try someplace new . \\ I searched online for barbers and salons that were highly rated in the area . I finally decided on one that \\ was fairly close by , highly reviewed , and had fair prices . I called the salon and made an appointment \\ for the next day . Upon arriving , I was greeted by a very nice hair stylist .... I was very happy with the \\ haircut . I then paid and gave her a tip for her service .
\vspace{1mm}
\\ Who cut their hair ?\\ The barber\\ The hair stylist\end{tabular}                                                                                                                                                                                                                     &  Dependency\_parse                      \\ \hline
\begin{tabular}[c]{@{}l@{}}I was at home watching the baby while my wife was at work . I feed the baby and was just watching TV \\ when I could tell that it was time to change the babies diaper . I went and got the diaper bag with every- \\ thing that I would need to change the baby . I laid the baby down on the changing table and removed the \\ old diaper and wiped down the baby to get her clean . Then I put some fresh powder on the baby and took \\ out a new fresh diaper and put it on the baby . Now the baby was all fresh and clean and smelled really nice \\ also . The baby was in a better mood after I changed her she was more at ease and happy because she was \\ clean and feeling better . I always try to stay on top of changing the babies diaper because they feel better \\ and it is better for them .
\vspace{1mm}
\\ Who changed the baby diaper ?\\ the mother\\ the father\end{tabular}                                                                                                                                                                                                                                                  & Co-reference                 \\ \hline

\end{tabular}
\end{table*}

\section{Future Work}
\begin{itemize}[leftmargin=7.5pt]
\item Lack of commonsense reasoning ability remains as the major cause of errors made by the network. We saw a significant improvement in the results by pre-training the PC and QC streams on auxiliary tasks.  Extending this, we might benefit from pre-training our central PQC network on other commonsense reasoning tasks like the Children's book test from Facebook bAbI.

\item We can compliment ConceptNet's loosely structured but large-scale commonsense knowledge with script based commonsense knowledge.  One way of inducing script based commonsense knowledge into the system is to pre-train the central PQC network on commonsense script knowledge like DeScript in the auto-encoder fashion. The idea is to implicitly encode the requisite background knowledge into the models parameters.

\begin{figure}[ht]
	\centering
   \includegraphics[width=0.5\textwidth]{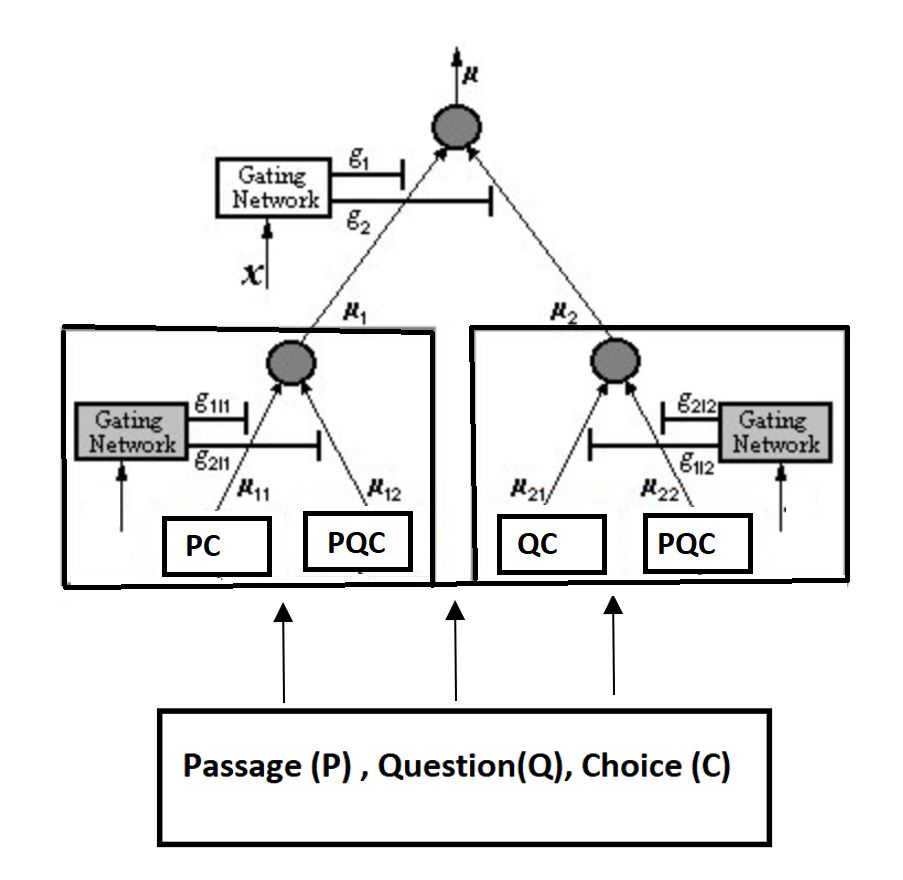}
   \caption{Future Work : Hierarchical mixture of experts }
   \label{fig: accuracy curve}
\end{figure}

\item Many errors are caused because of the network's lack of ability to interpret the co-references and subject-predicate dependencies in the passage. Including additional features in the passage embedding based on the dependency parse of the passage and using co-reference tags will surely help in improving the results.

\item Taking inspiration from the hierarchical mixture of experts, we might benefit from inducing decision layers that can jointly use the decision made by the underlying experts. In our case, combining the prediction from (PQC,PC) streams and parallely combining the decision from the (PC,QC) streams  and the (PQC,QC) streams would give rise to an architecture similar to what is shown in Figure 4. We believe this might give us a sophisticated way of combining the different classification results and improve the overall accuracy.
\end{itemize}

% \pagebreak
\bibliographystyle{ACM-Reference-Format}
\bibliography{main}
\end{document}